\documentclass[conference]{IEEEtran}
\IEEEoverridecommandlockouts

\usepackage{cite}
\usepackage{amsmath,amssymb}
\usepackage{graphicx}
\usepackage{booktabs}
\usepackage{hyperref}
\usepackage{url}
\usepackage{listings}
\usepackage{algorithm}
\usepackage{algpseudocode}
\title{When Corrective Hints Hurt: Prompt Design in\\
Reasoner-Guided Repair of LLM Overcaution\\
on Entailed Negations under OWL~2~DL}

\author{
\IEEEauthorblockN{Yijiashun Qi\textsuperscript{*}}
\IEEEauthorblockA{\textit{University of Michigan}\\
Ann Arbor, MI, USA\\
elijahqi@umich.edu}
\and
\IEEEauthorblockN{Xiang Xu}
\IEEEauthorblockA{\textit{ByteDance Inc.}\\
San Jose, CA, USA\\
xuxiang.victor@bytedance.com}
\and
\IEEEauthorblockN{Yuxuan Li}
\IEEEauthorblockA{\textit{University of Pennsylvania}\\
Philadelphia, PA, USA\\
yuxuanli@alumni.upenn.edu}
\thanks{\textsuperscript{*}Corresponding author.}
}

\begin{document}
\maketitle

\begin{abstract}
We report a reproducible error pattern in GPT-5.4 on OWL~2~DL compliance
queries: the model frequently answers ``unknown'' when the reasoner-entailed
answer is ``no'' under \emph{FunctionalProperty} closure or class
\emph{disjointness}.  Using 180 reasoner-audited queries from a procedural
expansion of the observed pattern plus 18 hand-authored held-out queries in
two unrelated domains (insurance and clinical), we compare four interaction
modes under matched query budget: single-shot, three rounds of generic
``you-are-wrong'' retry, three rounds of reasoner-verdict repair with an
open-world-assumption (OWA) hint, and the same repair without the hint.
Direct faithfulness is 43.9\,\% (Wilson 95\,\% CI $[36.8,51.2]$); generic
retry reaches 81.7\,\% ($[75.4,86.6]$); the verdict-with-hint variant is
\emph{worse} at 67.2\,\% ($[60.1,73.7]$); the verdict-only variant reaches
97.8\,\% ($[94.4,99.1]$).  All pairwise comparisons remain significant under
McNemar's exact test with Bonferroni correction
($\alpha = 0.01$; all $p < 10^{-5}$).  The same fingerprint accounts for 4/4
errors on the held-out queries.  Our interpretation is bounded: prompt
framing can matter more than corrective content, and reasoner-guided
wrappers should be ablated explicitly.
\end{abstract}


\section{Introduction}

LLMs are increasingly deployed alongside formal ontologies for
compliance-sensitive decision making.  In such settings, one-shot accuracy
is not enough: we also need to know whether an initially incorrect answer
can be \emph{corrected} under bounded verifier feedback.  We operationalise
this second axis --- \emph{auditability} --- as the conditional probability
that a wrong answer becomes right after $\leq k$ rounds of feedback, and we
treat it as distinct from single-shot faithfulness.

This paper presents a controlled empirical study of one specific failure
mode of GPT-5.4 on OWL~2~DL queries, and of three feedback regimes designed
to repair it.  Three findings drive our contribution.  \textbf{(1)}~GPT-5.4
systematically returns ``unknown'' rather than the entailed ``no'' when the
proof requires \emph{FunctionalProperty} closure or class \emph{disjointness};
$99/100$ direct-mode errors on our 180-query benchmark share this
fingerprint, and the same fingerprint explains $4/4$ errors on two
hand-authored held-out scenarios in different domains.  \textbf{(2)}~Three
rounds of generic retry (``your answer is wrong, try again'') raise
faithfulness from $43.9\%$ to $81.7\%$ ($p < 10^{-20}$), while three rounds
of reasoner-verdict repair with no semantic hint reach $97.8\%$.
\textbf{(3)}~A reasoner-verdict template that additionally includes the
logically-correct hint ``consider OWA --- missing triples are not
negations'' is $14.4$\,pp \emph{worse} than generic retry
($p = 1.3 \times 10^{-5}$).  The hint is associated with more ``unknown''
responses, consistent with over-application of OWA cues by the model.  The
resulting methodological point is the paper's main contribution:
\emph{in reasoner--LLM hybrids, prompt framing can matter more than the
corrective content it carries.}

\section{Background and Related Work}

Two OWL~2 features create entailed negations under an otherwise open-world
semantics: (i)~\emph{FunctionalProperty}, which fixes the asserted value as
the unique value of the property and closes the world for that property,
and (ii)~class \emph{disjointness}.  Combined with \emph{equivalent-class}
definitions, these let a sound reasoner derive $x \notin C$ without any
axiom directly asserting the negation.

Recent benchmarks evaluate LLMs on description-logic and ontology-grounded
tasks at varying scale: Wang \emph{et al.}~\cite{wang2024dllite} on
DL-Lite; DL-ReasonSuite~\cite{dlreasonsuite2026} at the 4k-task scale;
OntoURL~\cite{ontourl2025} with tens of thousands of questions;
Yang \emph{et al.}~\cite{yang2026owlproofs} on natural-language proofs.
Systematic LLM benchmarking has also spread across modalities and domains,
e.g.\ Qi \emph{et al.}~\cite{qi2024marine} on domain-specific image
classification, Qi \emph{et al.}~\cite{qi2026econagents} on simulating
firm-level economic behaviour, and Yang \emph{et al.}~\cite{yang2025tinybert}
on distilling semantic knowledge into smaller language models.
Complementing these, Xing \emph{et al.}~\cite{xing2025structuredmemory}
study structured memory mechanisms for stable context representation in
LLMs at ICAIDE 2025; we build on that line of work by focusing on a
specific reasoning failure rather than general context stability.
Knowledge structures also interact with LLMs on the retrieval side:
Qi \emph{et al.}~\cite{qi2026webpipeline} instantiate a
web--knowledge--web pipeline for domain discovery, and graph neural
networks have been applied to structured business-data reasoning in
\cite{qi2025gnnmining, qi2026hetgnnsme}.  Compared to these, our work is a
\emph{targeted stress test}: we trade scale for tight per-query auditing
and a focused study of one structural failure pattern.  Within the
reasoner--LLM hybrid literature, Explanation-Refiner~\cite{explanationrefiner2024}
uses verifier-derived feedback to iteratively refine LLM outputs.  Chain-of-thought
prompting~\cite{wei2022chain} shows that surface cues in the prompt can alter
the reasoning path a model traverses; the phenomenon we report can be read as
a failure mode of that same mechanism.  We isolate a previously unreported
\emph{prompt-wrapping} effect inside this class of systems.

\section{Method}

\subsection{Scenarios}
We use three sets.  The \textbf{development set} comprises 10 hand-authored
mini-ontologies (subsumption, numeric boundaries, OWA traps, one
mixed-compliance case; 38 queries).  The \textbf{expansion set} consists of
30 procedurally generated variants of the mixed-compliance template,
parameterised over a Cartesian grid of spend buckets (below threshold, at
strict boundary, above) and class memberships (active, blacklisted, none);
180 queries.  The \textbf{held-out set} contains two further hand-authored
ontologies in insurance and clinical domains with disjoint vocabularies;
18 queries.  Listing~\ref{lst:example} shows one representative generated
variant.  The scenario combines a \texttt{FunctionalProperty}
(\texttt{hasSpend}), an equivalent-class definition
(\texttt{VIPCustomer}), disjoint classes
(\texttt{ActiveCustomer}/\texttt{Blacklisted}), and a second
equivalent-class definition chaining the two
(\texttt{ActiveVIP}).  A query such as ``Is \texttt{c\_a} an
\texttt{ActiveVIP}?'' has the reasoner-entailed answer ``no'' because the
asserted functional value $500 \not> 1000$; yet GPT-5.4 in direct mode
returns ``unknown''.

\begin{figure*}[t]
\begin{lstlisting}[caption={A representative generated scenario
(abridged Turtle).  The correct answer to ``Is \texttt{c\_a} an
\texttt{ActiveVIP}?'' is ``no'' via FunctionalProperty closure;
GPT-5.4 in direct mode returns ``unknown''.},
label={lst:example}]
:hasSpend      a owl:FunctionalProperty, owl:DatatypeProperty ;
               rdfs:range xsd:decimal .

:VIPCustomer   owl:equivalentClass [
                 a owl:Class ;
                 owl:intersectionOf (
                   :Customer
                   [ owl:onProperty :hasSpend ;
                     owl:someValuesFrom [
                       a rdfs:Datatype ;
                       owl:onDatatype xsd:decimal ;
                       owl:withRestrictions ( [ xsd:minExclusive 1000 ] ) ] ] ) ] .

:ActiveVIP     owl:equivalentClass [ owl:intersectionOf ( :VIPCustomer :ActiveCustomer ) ] .

[ a owl:AllDisjointClasses ; owl:members ( :ActiveCustomer :Blacklisted ) ] .

:c_a a :Customer ; :hasSpend 500  ; a :ActiveCustomer .
:c_b a :Customer ; :hasSpend 1500 ; a :ActiveCustomer .
\end{lstlisting}
\end{figure*}

\subsection{Ground truth}
Every query's gold answer is computed by a sound reasoner
(HermiT~\cite{glimm2014hermit} by default; Pellet~\cite{sirin2007pellet}
when JDK~$\geq 21$ is configured) via the consistency-based
trial-insertion protocol of Algorithm~\ref{alg:tri}.  The procedure
distinguishes entailed negation from open-world unknown by running the
reasoner twice --- once on the KB augmented with the positive claim, once
on the KB augmented with the negated claim --- and interpreting any
resulting inconsistency as entailment of the opposite.  Every authored
answer matches the reasoner's verdict.

\begin{algorithm}[t]
\caption{Consistency-based trial insertion for three-valued entailment.}
\label{alg:tri}
\begin{algorithmic}[1]
\Require Ontology $\mathcal{O}$, query $x \in C$
\Ensure Answer $\in \{\text{yes}, \text{no}, \text{unknown}\}$
\State $\mathcal{O}^{+} \gets \mathcal{O} \cup \{C(x)\}$
\State $\mathcal{O}^{-} \gets \mathcal{O} \cup \{\neg C(x)\}$
\State $\textit{pos} \gets \textsc{Consistent}(\mathcal{O}^{+})$
\State $\textit{neg} \gets \textsc{Consistent}(\mathcal{O}^{-})$
\If{$\textit{pos} \wedge \neg\textit{neg}$} \Return yes
\ElsIf{$\neg\textit{pos} \wedge \textit{neg}$} \Return no
\ElsIf{$\textit{pos} \wedge \textit{neg}$} \Return unknown
\Else{} \Return \textbf{error} (KB already inconsistent)
\EndIf
\end{algorithmic}
\end{algorithm}

\subsection{Models and modes}
The repair-design analysis focuses on GPT-5.4, accessed via the OpenAI
Responses-protocol endpoint of an internal API gateway during April~2026
at default decoding settings.  We compare four modes under a matched budget
(one prompt + up to three follow-ups):
\textbf{direct} (single-shot JSON answer);
\textbf{na\"ive\_repair} (``Your answer is incorrect.  Reconsider.'');
\textbf{cx\_repair v1} (same, plus the reasoner's verdict \emph{and} the
hint ``consider OWA: missing triples are not negations'');
\textbf{cx\_repair v3} (reasoner's verdict only, no hint).
For cross-model context only, we also evaluate GLM-5, Kimi-k2.5 and
Minimax-2.7 on the development set via the same gateway
(Anthropic-style protocol); these are descriptive, not controlled.

\subsection{Metrics}
Faithfulness is the fraction of queries whose final answer matches the
reasoner verdict.  Confidence intervals use Wilson's score interval
($\alpha = 0.05$).  Pairwise mode comparisons use McNemar's exact two-sided
test on discordant pairs.  We apply Bonferroni correction for five tests
(family-wise $\alpha = 0.01$); all five reported tests remain significant
($p \leq 1.3 \times 10^{-5}$).

\section{Results}

\subsection{Main result: 180-query expansion set}

\begin{figure}[t]
\centering
\includegraphics[width=0.92\columnwidth]{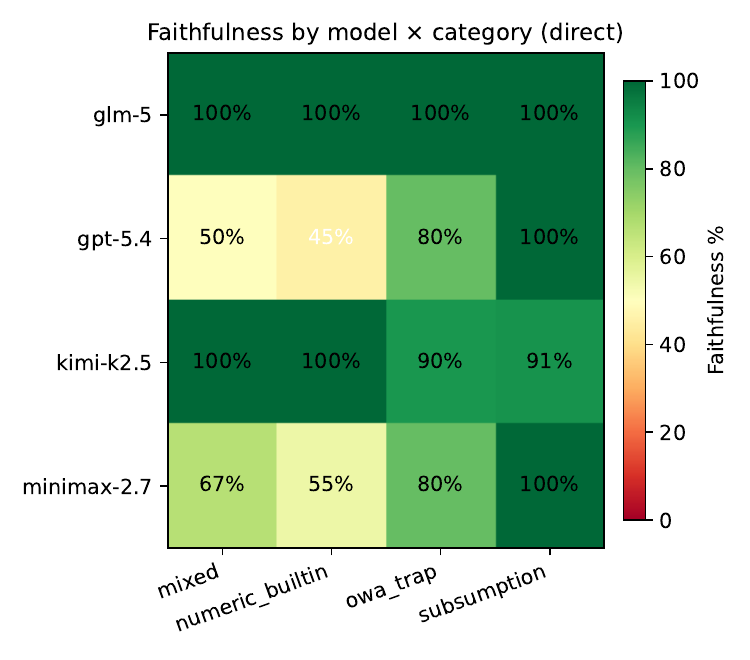}\\[-0.1em]
\includegraphics[width=0.92\columnwidth]{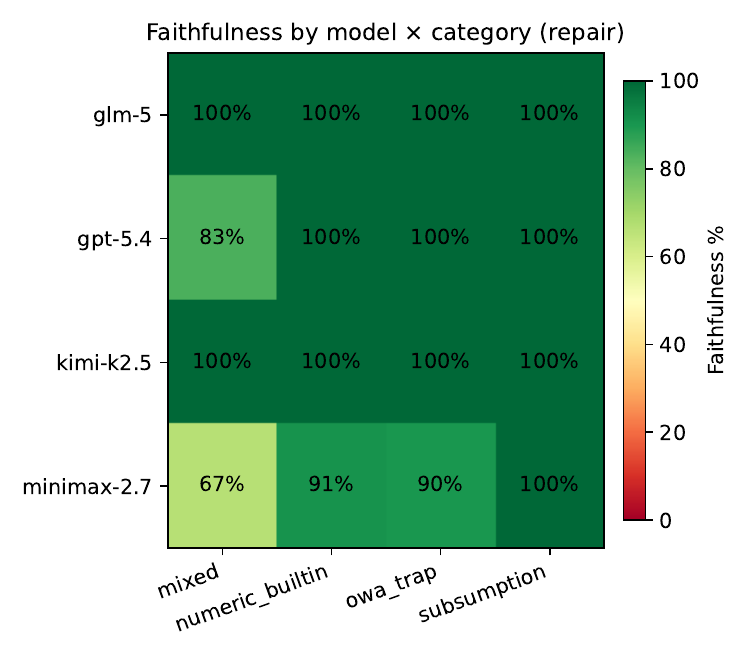}
\caption{Per-category faithfulness on the development set.
\textbf{Top:} single-shot direct mode --- GPT-5.4's deficit concentrates
in \emph{mixed} (FunctionalProperty closure + disjointness) and
\emph{numeric\_builtin}.
\textbf{Bottom:} after counterexample-driven repair --- GPT-5.4 closes
most of the gap; residual errors concentrate in the mixed category.}
\label{fig:heatmap}
\end{figure}

Table~\ref{tab:main} gives GPT-5.4's faithfulness across the four modes.
Figure~\ref{fig:heatmap} visualises per-category performance on the
development set, directly and after repair.

\begin{table}[t]
\caption{GPT-5.4 on the 180-query expansion set: Wilson 95\,\% CIs and
pairwise McNemar tests (Bonferroni-corrected $\alpha = 0.01$).}
\label{tab:main}
\centering
\scriptsize
\begin{tabular}{lrr}
\toprule
Mode & Faithfulness & 95\,\% CI \\
\midrule
direct                     & 79/180 = 43.9\,\%  & $[36.8, 51.2]$ \\
na\"ive\_repair             & 147/180 = 81.7\,\% & $[75.4, 86.6]$ \\
cx\_repair (hint)          & 121/180 = 67.2\,\% & $[60.1, 73.7]$ \\
cx\_repair (verdict-only)  & \textbf{176/180 = 97.8\,\%} & $[94.4, 99.1]$ \\
\midrule
\multicolumn{3}{l}{\emph{Pairwise McNemar, paired}} \\
\midrule
direct $\to$ na\"ive         & $\Delta = +37.8$\,pp & $p = 6.8 \times 10^{-21}$ \\
direct $\to$ cx\_v3          & $\Delta = +53.9$\,pp & $p = 1.3 \times 10^{-29}$ \\
na\"ive $\to$ cx\_v1 (hint)  & $\Delta = -14.4$\,pp & $p = 1.3 \times 10^{-5}$ \\
na\"ive $\to$ cx\_v3         & $\Delta = +16.1$\,pp & $p = 1.3 \times 10^{-7}$ \\
cx\_v1 $\to$ cx\_v3          & $\Delta = +30.6$\,pp & $p = 8.0 \times 10^{-16}$ \\
\bottomrule
\end{tabular}
\end{table}

\begin{figure}[t]
\centering
\includegraphics[width=0.92\columnwidth]{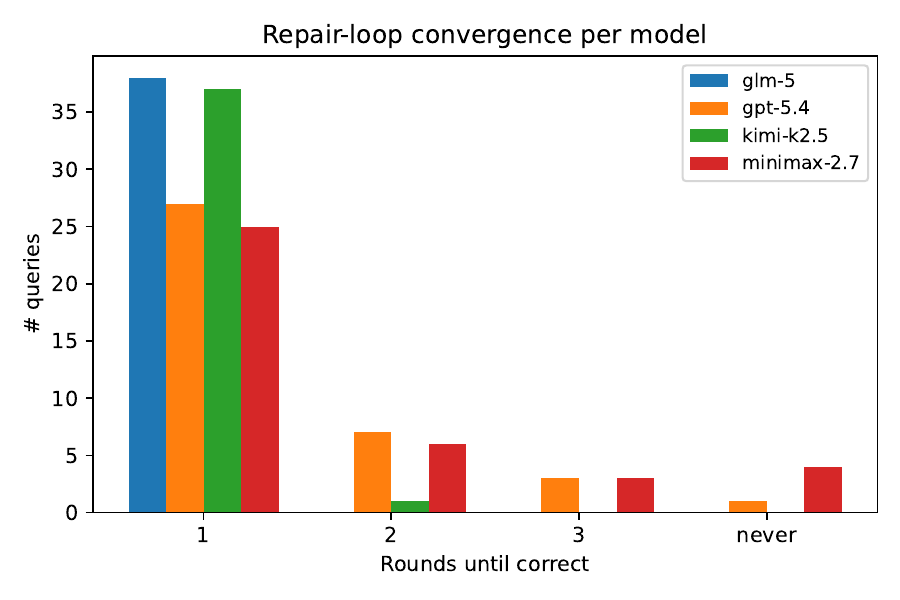}
\caption{Rounds-until-correct per model on the development set under
counterexample repair (max 3 rounds).  GPT-5.4 requires more rounds than
the other three models; the ``never'' bar marks queries still wrong after
the three-round budget.}
\label{fig:convergence}
\end{figure}

\begin{figure}[t]
\centering
\includegraphics[width=0.92\columnwidth]{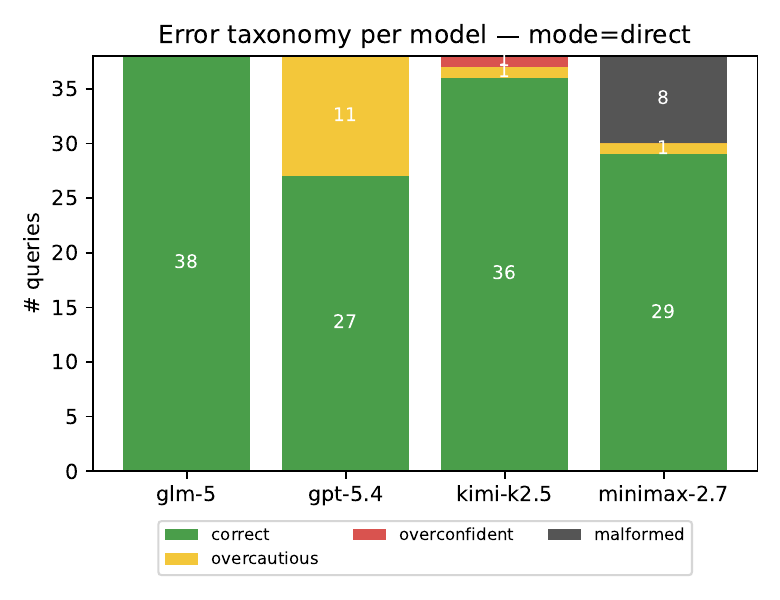}
\caption{Direct-mode error taxonomy on the development set.  GPT-5.4's
errors are dominated by \emph{overcautious} (``unknown'' where the reasoner
says ``no''); Minimax's errors are dominated by \emph{malformed} JSON
(a reasoning-model token-budget artefact rather than a semantic error).}
\label{fig:taxonomy}
\end{figure}

\subsection{Worked example: one query, two wrappers}

Listing~\ref{lst:transcripts} juxtaposes the \emph{verdict-only} and
\emph{verdict-with-hint} follow-up prompts applied to the same failing
query on the scenario of Listing~\ref{lst:example}.  The two prompts
deliver the same truth (``the correct answer is no'') with different
surface phrasing.  Under the verdict-only prompt, GPT-5.4 revises to
``no'' within two rounds; under the hint-bearing prompt, it re-emits
``unknown'' and cites OWA as justification --- reproducing the failure
the reasoner has just flagged.  This anecdotal reproduction of the
aggregate effect helps explain why the hint may be pulling the model
toward ``unknown'' rather than away from it.

\begin{lstlisting}[caption={Follow-up prompts (shortened) after a wrong
``unknown'' answer, and the next-round GPT-5.4 response. Only the wrapper
differs.}, label={lst:transcripts}]
-- VERDICT-ONLY WRAPPER --
[system]  The OWL 2 DL reasoner verified the correct answer is "no", not "unknown".
[GPT-5.4] {"answer":"no","reason":"hasSpend is Functional and its value 500 is not > 1000,
           so VIPCustomer is not satisfied, hence not ActiveVIP."}

-- VERDICT-WITH-HINT WRAPPER --
[system]  The OWL 2 DL reasoner verified the correct answer is "no", not "unknown".
          Consider OWA: missing triples are not negations.
[GPT-5.4] {"answer":"unknown","reason":"Under OWA we cannot conclude the negation of
           ActiveVIP from the absence of a triple."}
\end{lstlisting}

\subsection{Failure-mode characterisation}

Of the 101 errors GPT-5.4 makes in direct mode on the expansion set,
\textbf{99} share the exact pattern ``expected\,=\,no, got\,=\,unknown''.
Every one of these errors involves either FunctionalProperty closure
(the asserted value provably satisfies a strict numeric relation against
the threshold) or class disjointness.  No errors are inversions
(yes$\to$no or no$\to$yes); only one is malformed JSON.  The failure is
therefore not random hallucination but \emph{structural overcaution
against entailed negation}.

\subsection{Held-out and cross-model context}

On the 18 held-out queries (two hand-authored ontologies in insurance and
clinical domains), direct mode reaches 14/18 (77.8\,\%); the four errors
are all ``expected\,=\,no, got\,=\,unknown'', confirming that the failure
pattern is not a procedural-template artefact.  Both repair variants reach
18/18 on the held-out set, but $n = 18$ is too small to separate them.
On the development set, three other frontier models from the same
gateway --- GLM-5, Kimi-k2.5, Minimax-2.7 --- do not exhibit the
overcaution pattern in direct mode (38/38, 36/38, 29/30 semantic
faithfulness respectively).  We report these only to rule out a trivial
``all LLMs share this overcaution'' explanation; because the four models
differ in protocol and default decoding, the comparison is not controlled.

\section{Discussion}

\subsection{Why the OWA hint backfires: a qualitative analysis}
The hint we added to cx\_repair v1 --- ``consider OWA: missing triples
are not negations'' --- is logically correct and is exactly the kind of
guidance a human ontology engineer would give a learner confusing
closed- and open-world reasoning.  Yet on this failure mode (entailed
negation \emph{via} closure) the hint coincides with a measurable shift
toward ``unknown'' precisely where the correct answer is ``no''.

To probe \emph{how} the hint redirects the model, we inspected the 59
errors cx\_v1 makes on the expansion set against the 4 errors cx\_v3
makes on the same queries.  Of the 59 cx\_v1 errors, \textbf{56} share
the exact ``expected\,=\,no, got\,=\,unknown'' fingerprint despite three
rounds of feedback stating the reasoner-verified correct answer is
``no''; the remaining 3 are malformed JSON.  We read the transcripts
round-by-round and observed a consistent pathway: after receiving the
OWA hint, the model frequently emits reasoning that \emph{acknowledges}
FunctionalProperty closure (``\texttt{hasSpend} is functional, so
\texttt{c\_a} can have at most one spend value'') yet still concludes
``unknown''.  One representative trace on \texttt{gen24.c\_a\_24.vip}
reads:

\begin{quote}
\small
``Since \texttt{hasSpend} is functional and \texttt{c\_a} has
\texttt{hasSpend 500.0}, \texttt{c\_a} cannot also have a different
spend value $>1000.0$; thus \texttt{c\_a} cannot satisfy the existential
restriction for \texttt{VIPCustomer}.  Therefore \texttt{c\_a} is not a
\texttt{VIPCustomer}.''
\end{quote}

\noindent --- and then the model still outputs \texttt{"unknown"}.  The
chain of reasoning points to a definite negation while the final label
does not follow.  Our interpretation is \emph{cue over-application}:
the OWA hint activates a surface hedging heuristic that dominates the
model's own closure-based proof, consistent with chain-of-thought work
showing that surface cues can override the reasoning
content~\cite{wei2022chain}.  We present this as a plausible mechanism
rather than a proven one: with only API access we cannot inspect the
hidden state; a direct-weights experiment would be needed to confirm
the pathway.  The verdict-only template, which omits the OWA cue,
avoids this shift in all but 4 cases.

\subsection{Computational cost and latency}
Table~\ref{tab:latency} reports per-query latency and total wall-clock
time for each mode on the 180-query expansion set.  All modes ran
through the same gateway under identical network conditions, so the
numbers are directly comparable within this study (absolute latencies
on a direct provider API may differ).

\begin{table}[t]
\caption{Latency and query budget per mode (180 queries, GPT-5.4).}
\label{tab:latency}
\centering
\scriptsize
\begin{tabular}{lrrr}
\toprule
Mode & Avg.\ latency & Avg.\ rounds & Total wall-clock \\
\midrule
direct                       & 4.07\,s & 1.00 & 12.2\,min \\
na\"ive\_repair               & 4.90\,s & 1.76 & 25.8\,min \\
cx\_repair (hint)            & 5.05\,s & 1.93 & 29.3\,min \\
cx\_repair (verdict-only)    & \textbf{4.77\,s} & \textbf{1.63} & \textbf{23.3\,min} \\
\bottomrule
\end{tabular}
\end{table}

Two observations follow.  First, the repair modes roughly double the
wall-clock cost of single-shot prompting, so the strategy is most
naturally deployed in batch or offline compliance validation rather
than in the hot path of an interactive user loop.  For interactive use,
the reasoner audit can be gated behind confidence filters or invoked
only for cases the reasoner would otherwise resolve as ``no''.
Second, among the repair modes, the verdict-only variant is not only
the most accurate (97.8\,\% vs 67.2\,\% for the hint variant) but also
the \emph{cheapest}: 1.63 average rounds vs 1.93, a 16\,\% reduction in
repair cost alongside a 30.6\,pp accuracy gain.  Counter-intuitively,
the richer prompt is strictly dominated on both axes.

\subsection{Implications for reasoner--LLM hybrids}
A natural assumption in reasoner--LLM hybrid systems is that more
reasoner output $\Rightarrow$ better LLM correction.  Our data refute
this: the richer counterexample template is associated with a 30.6-pp
reduction in repair quality compared to the verdict-only variant
(Table~\ref{tab:main}, $p = 8 \times 10^{-16}$).  For practitioners
building reasoner-in-the-loop systems, three implications follow.
First, the wrapper around the reasoner's verdict should be treated as a
\emph{hyperparameter} and ablated explicitly; any new prompt template
should be benchmarked against a content-light baseline on a targeted
failure set.  Second, \emph{explanations that are logically correct but
topically mismatched} can be worse than no explanation at all; the
generic advice to ``teach the model why it was wrong'' needs a
mode-specific refinement.  Third, structural overcaution on entailed
negation, if unaddressed, constrains the range of ontological
constructs that can safely be put behind an LLM interface in a
compliance pipeline --- rules that rely on FunctionalProperty closure
or class disjointness are exactly the ones a human compliance engineer
would write, and exactly the ones our data show the model struggles to
honour unaided.

\section{Ethics, Reproducibility, Threats, and Conclusion}

\subsection{Ethics and digital-ethics relevance}

Compliance-sensitive deployments of LLMs (e.g.\ credit, insurance, or
clinical triage) often rest on rules that are naturally expressible as
ontological closures --- a functional attribute bounds a quantity, a
disjointness axiom bounds a category.  A model that systematically
returns ``unknown'' where the rule logically entails ``no'' is not
harmlessly cautious: downstream it may degrade to either silent
abstention or unprincipled over-deferral to a human reviewer, neither of
which is auditable in the sense of~\cite{xing2025structuredmemory}.
Our study isolates such a failure mode on a narrow but realistic pattern
and shows it is correctable through bounded feedback.  The finding that
a logically-correct but mode-mismatched hint can \emph{degrade} a
reasoner-guided loop has direct ethical implications for practitioners
wiring formal verifiers into agentic pipelines: verifier feedback that
looks safer on a surface read may in fact bias the model toward the
exact behaviour the reviewer would flag.

\subsection{Reproducibility}

All 10 development scenarios, the 30-variant generator, both held-out
ontologies, and every raw model transcript are preserved as JSON for
every run reported.  Reasoner gold answers are computed by
Algorithm~\ref{alg:tri} and verified to match hand-authored labels on
all 38 development queries.  We will release the harness, scenario
generator, and results upon publication.  The entire pipeline ---
scenario construction, reasoner audit, LLM querying, statistical
analysis --- runs from a single command on commodity hardware with a
CPython~3.13 interpreter, a Java runtime (11 for HermiT, 21 for
Pellet), and network access to a provider-specific LLM endpoint.

\subsection{Future work}

Three directions follow directly from the present stress test.
First, a \emph{cross-family} failure survey that extends the
auditability framework (direct vs na\"ive retry vs verdict vs
verdict-with-hint) across Western frontier models --- Claude~4.x,
Gemini~2.x, Llama~3 --- as well as additional release dates of GPT
itself.  Replicating the fingerprint (or ruling it out) on these
families is the single highest-value follow-up, and separates
model-specific pathologies from a broader architectural tendency of
autoregressive LLMs to hedge on entailed negations.  Second, a
systematic sweep over hint vocabularies: the single hint we ablated is
one point in a large design space of reasoner-feedback templates;
minimal-pair controls across synonyms, negation markers, and
rule-grounding phrases could map where the backfire region begins and
ends, and whether the effect survives paraphrase.  Third, composing
auditability-aware repair with retrieval-augmented or graph-augmented
inference~\cite{qi2025gnnmining, qi2026hetgnnsme, qi2026webpipeline}:
if the primary error source is structural overcaution, a verifier
signal that locates the \emph{triggering axiom} (rather than merely
reporting a verdict) may close the residual gap visible in
Fig.~\ref{fig:heatmap}.  A parallel line of inquiry is scaling the
evaluation to large, publicly available OWL~2~DL ontologies --- see
the next subsection for why our targeted-stress-test design could not
absorb those outright.

\subsection{Threats to validity}

This study has limits.  \textbf{Single model under detailed study.}
Our deep characterisation is of GPT-5.4 only.  The cross-model context
(Sec.~V-D) shows three other frontier models accessed through the same
gateway (GLM-5, Kimi-k2.5, Minimax-2.7) do not exhibit the overcaution
fingerprint on the development set, which rules out a trivial ``all
LLMs share this overcaution'' reading; it does not establish that
Western frontier models (Claude, Gemini, Llama~3) are immune.  We did
not have stable, controlled access to those endpoints during the
study period and defer the full cross-family survey to future work.
Our claim is correspondingly narrow --- a reproducible failure mode
on one deployed model --- not a universal architectural diagnosis.

\textbf{Templated expansion.}  The 180-query main set is generated
from one Cartesian template, but the held-out scenarios in two
unrelated domains (insurance, clinical) preserve the same failure
fingerprint, showing that the effect is not a procedural-template
artefact.

\textbf{Scale and ontology complexity.}  This is a targeted stress
test, not a general ontology-reasoning benchmark.  Our 180\,+\,18\,+\,38
queries are dwarfed by DL-ReasonSuite~\cite{dlreasonsuite2026} and
OntoURL~\cite{ontourl2025}.  A natural question is whether the
verdict-only repair strategy holds on large production ontologies such
as SNOMED~CT or the Gene Ontology.  We flag two considerations.
(i)~SNOMED~CT and the Gene Ontology are authored in the OWL~2~EL
profile, which \emph{excludes} \texttt{FunctionalProperty} and
\texttt{DisjointClasses} axioms for tractability; the exact closure
mechanisms that trigger the fingerprint we study therefore do not
arise in pure EL ontologies, so the failure pattern is not directly
reproducible on them.  (ii)~Larger OWL~2~DL ontologies that
\emph{do} include these constructs --- e.g.\ the NCI Thesaurus, GALEN,
or BFO-grounded extensions of SNOMED --- are the natural next
evaluation target, and we expect the prompt-wrapping effect to
persist structurally while the repair budget and reasoner-audit cost
both grow with reasoning depth.  A controlled scaling study on these
ontologies is deferred to future work.

\textbf{Closed-model determinism.}  We accessed all models via a
third-party gateway in April~2026 with no client-side control over
decoding temperature or random seed, and did not sweep prompt
paraphrases; replication under perturbation and on direct provider
APIs is deferred.  Provider-side updates may also change the model's
behaviour without a public version bump.

\textbf{Held-out sample size.}  The 18-query held-out set is too small
to power pairwise mode comparisons; it is intended only to test
whether the direct-mode failure pattern transfers to vocabulary
unrelated to the expansion template.

\subsection{Conclusion}

We characterise a reproducible structural failure of GPT-5.4 on OWL~2~DL
queries that require FunctionalProperty closure or class disjointness ---
``overcaution against entailed negation''.  Generic retry recovers most
of the deficit; reasoner-guided repair with a content-light verdict
template recovers nearly all of it; reasoner-guided repair with a
logically correct but mode-mismatched semantic hint recovers
\emph{less} than generic retry.  The wrapper around corrective feedback
can dominate its content.  We release the harness, scenario generators,
and reasoner-audited gold answers.

\bibliographystyle{IEEEtran}

\end{document}